\documentclass[smallextended]{svjour3}       
\smartqed  
\usepackage{graphicx}
\usepackage{algorithm}
\usepackage{algorithmic}
\usepackage{amssymb}
\usepackage{times}
\usepackage{epsfig}
\usepackage{graphicx}
\usepackage{amsmath}
\usepackage{amssymb}
\usepackage{caption}
\usepackage{subcaption}
\usepackage{amssymb}
\usepackage{pifont}
\usepackage{url}
\usepackage{pifont}
\usepackage{amssymb}
\usepackage[numbers]{natbib}
\usepackage[pagebackref=false,breaklinks=true,letterpaper=true,colorlinks,
bookmarks=false]{hyperref}
\usepackage{hyperref}

\usepackage{lineno}
\usepackage{multirow}
\usepackage{float}
\begin{document}

\title{%
	A Comprehensive Review on Sentiment Analysis: Tasks, Approaches and Applications 
	}


\author{Sudhanshu Kumar$^{*1}$ \and Partha Pratim Roy$^{1}$ \and Debi Prosad Dogra $^{2}$ \and Byung-Gyu Kim $^{3}$ 
}

\institute{Sudhanshu Kumar\\
	\email{skumar2@cs.iitr.ac.in}\\ 
	Partha Pratim Roy\\
	\email{partha@cs.iitr.ac.in}\\
	Debi Prosad Dogra \\  
	\email{dpdogra@iitbbs.ac.in}\\
	Byung-Gyu Kim\\
	\email{bg.kim@sookmyung.ac.kr}\\
	\\$^1$Department of Computer Science and Engineering, 
	IIT Roorkee, 247667, India\\
	$^2$School of Electrical Sciences, IIT Bhubaneswar, Odisha 752050, India. \\
	$^3$Department of IT Engineering, Sookmyung Women's University, Seoul 04310, South Korea \\
	\\$^*$Corresponding Author
	}

\date{Received: date / Accepted: date}

\maketitle

\begin{abstract}

Sentiment analysis (SA) is an emerging field in text mining. It is the process of computationally identifying and categorizing opinions expressed in a piece of text over different social media platforms. Social media plays an essential role in knowing the customer mindset towards a product, services, and the latest market trends. Most organizations depend on the customer's response and feedback to upgrade their offered products and services. SA or opinion mining seems to be a promising research area for various domains.  It plays a vital role in analyzing big data generated daily in structured and unstructured formats over the internet. This survey paper defines sentiment and its recent research and development in different domains, including voice, images, videos, and text. The challenges and opportunities of sentiment analysis are also discussed in the paper.

    	
\end{abstract}

\section{Introduction}\label{intro}
Sentiment Analysis is the computational study of people's opinions, attitudes, and emotions in the form of different modalities (text, image, and speech) toward an entity that represents topics, events, issues, products, services, and organizations. SA is the branch of many fields such as machine learning, data mining, natural language processing, and computational linguistics. Natural Language Processing (NLP) generally started back in the 1950s; little attention was paid by researchers to people's opinions and sentiment analysis until 2005. With the advancement of web 2.0, web 3.0, web 4.0, social media thrusts SA's development. Social media propels the growth of sentiment analysis.   
Most of the literature is on the English language, but many publications currently tackle the multilingual issue. SA is a suitcase research problem \cite{cambria2017sentiment} that is the combination of NLP tasks such as named entity recognition \citep{ma2016label}, concept extraction \citep{cambria2016senticnet},
sarcasm detection \citep{poria2016deeper}, aspect extraction \citep{ma2018targeted}, and subjectivity
detection. Subjective information indicates the opinions of opinion holders, while objective texts show some objective facts. For example, "The food is great and delicious." These opinion words are subjective. Subjective texts can have a positive or negative sentiment. 

\begin{figure}[!h!t!b]
	\begin{center}
		\includegraphics[width=1\columnwidth]{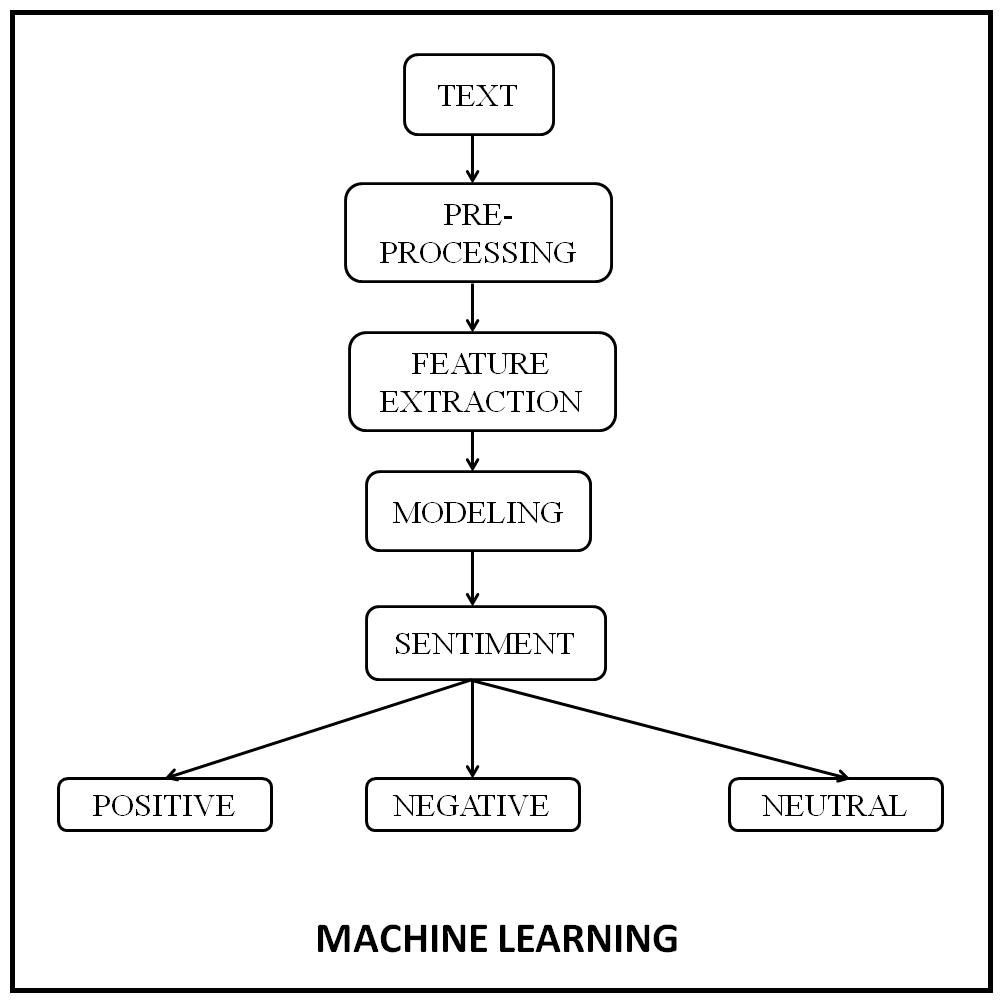}
		\caption{The process to classify the review into positive, negative, neutral using Machine learning.}
		\label{Data1}        
	\end{center}
\end{figure}

\begin{figure}[!h!t!b]
	\begin{center}
		\includegraphics[width=1\columnwidth]{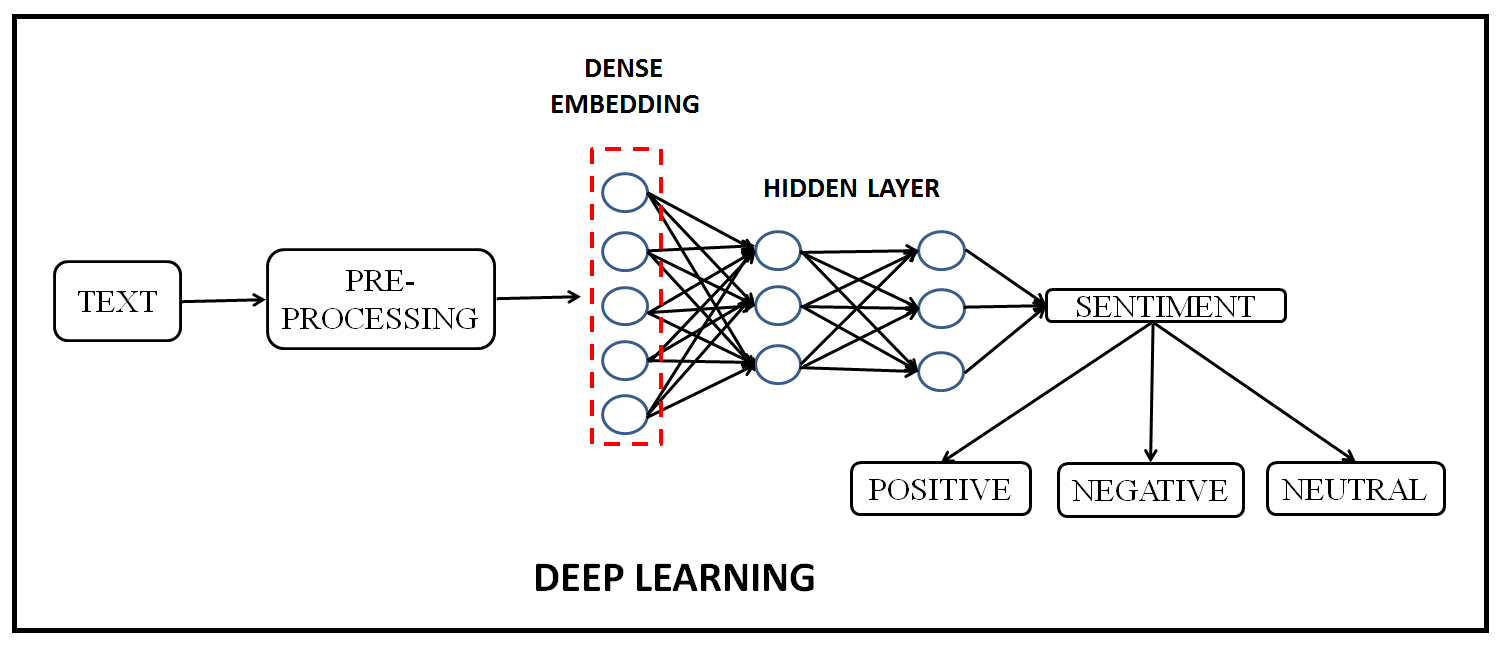}
		\caption{The process to classify the review into positive, negative and neutral using Deep learning techniques.}
		\label{Data111}        
	\end{center}
\end{figure}

SA classification process, as shown in Figure~\ref{Data1} and ~\ref{Data111} uses any classification model to classify the reviews into positive, negative and neutral classes. There are three levels of SA such as document level, sentence level, and aspect level. In the document level, the whole document expresses a positive or negative opinion. For Example, the product reviews document has either positive or negative opinions for a product.  It represents a single opinion for a document, so it comes under the document level.
Sentence level is the second category widely used in e-commerce sites in which each sentence classifies into positive, negative, and neutral opinions. Aspect level sentiment analysis is also called feature-based analysis. In this type of analysis, each review categorizes into aspects and their target opinions. This level shows more insights about the opinion that it is positive or negative for which aspect. For Example, ‘The Food was very good at the hotel.’ It is an aspect-based SA where food is one aspect of the review.


\begin{figure}[!h!t!b]
	\begin{center}
		\includegraphics[width=\linewidth]{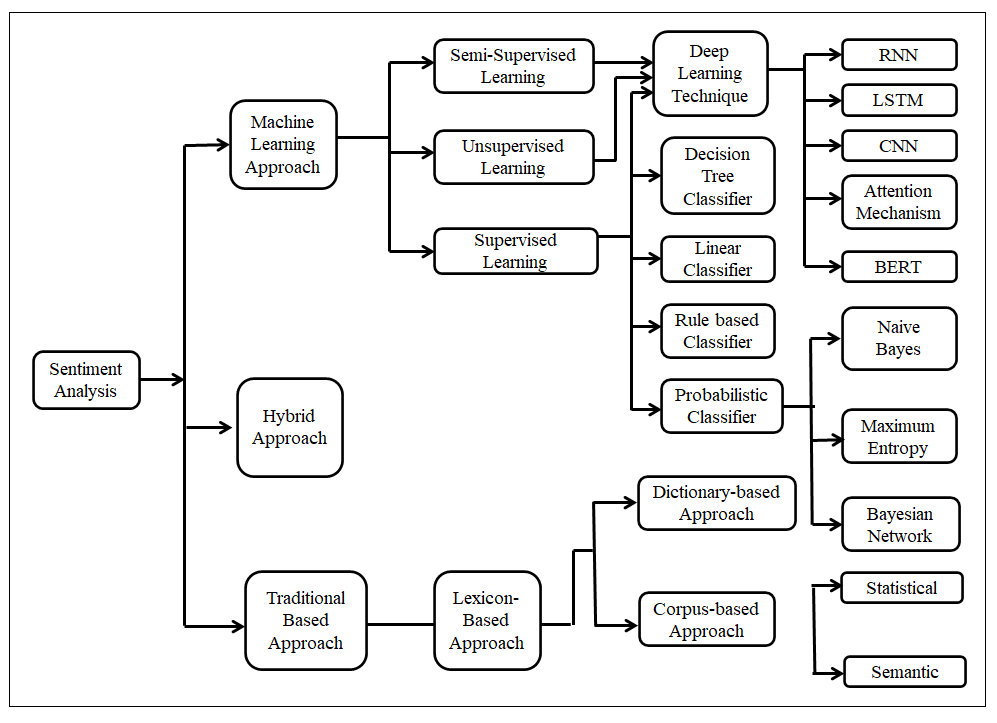}
		\caption{All sentiment analysis classification techniques from traditional to the latest one have been shown in this figure. Initially, It is divided into machine learning and lexicon-based approach, which further divide into different algorithms.}
		\label{Data4}        
	\end{center}
\end{figure}

\begin{figure}[]
	\begin{center}
		\includegraphics[width=1\linewidth]{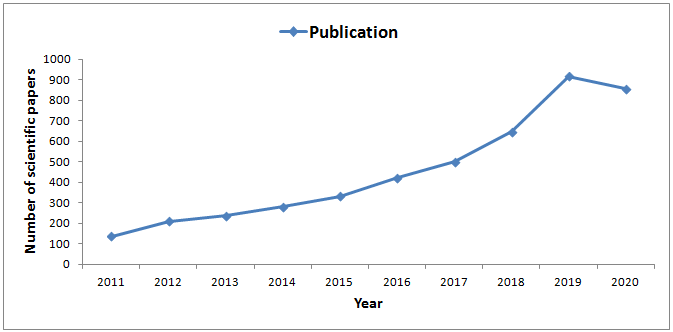}
		\caption{The number of research papers with the Scopus index are published in the last ten years from 2011 to Mid 2020. The number of publications is increasing yearly as the advancement in technology and the evolution of industry 4.0.}
		\label{Data6}        
	\end{center}
\end{figure}

Different sentiment classification techniques are shown in Figure~\ref{Data4}. It is divided into two categories, i.e., lexicon-based approach and machine learning approach. The Lexicon-based approach uses the dictionaries of words annotated with their semantic orientation, classified into the dictionary and corpus-based approach. The second category is the machine learning approach based on different types of learning like supervised, unsupervised, and semi-supervised. Depending on the nature of the data, these learning techniques are used and predict the result. Different deep learning-based and machine learning-based techniques are the most popular ones. The total number of research publications year-wise is as shown in Figure~\ref{Data6}. It shows that as the advancement of industry 4.0 and now it's 5.0, the numbers of research papers are increasing year by year. 


 The contributions of the paper are as follows:
\begin{enumerate}
	\item A large number of literature has been reviewed in sentiment analysis process from multiple domains and identify the pros and cons of all approaches.
	\item Summarizing each of the surveyed articles in detail, including the problems addressed, dataset details and methods.
       \item Analyses of existing applications in order to determine which one is most suitable for certain application. 
       \item Discussing the challenges and application of sentiment analysis in order to keep up the current research trends.  
\end{enumerate}

The rest of the paper is organized as follows. In Section \hyperref[sec:relwrk]{2}, we discuss the state of the art discussion on SA. Detailed discussion on the existing work, open issues, and possible applications of sentiment analysis is presented in Section \hyperref[sec:application]{3} and Section \hyperref[sec:challenges]{4}. Finally, in the last Section \hyperref[sec:conclude]{5} the research work has been concluded.

\section{Terminology and background concepts}
\label{sec:relwrk}
Opinion, views, and feeling are often used interchangeably in the different literature \cite{munezero2014they}. SA is also related to many terms such as emotions, moods, and feelings, which sometimes confuse the reader with opinion or SA. Emotion is related to the perception of the stimulus and the triggering of the bodily response. For example, one person shows an angry response when they lose the job, and in another case, they feel joy in the same situation. Many authors {\citep{munezero2014they,scherer2005emotions}} showed that emotion is short term, while the mood is a long-term phenomenon. They differentiate on both terms basis on their duration.

\begin{figure}[!h!t!b]
	\begin{center}
		\includegraphics[width=1\linewidth]{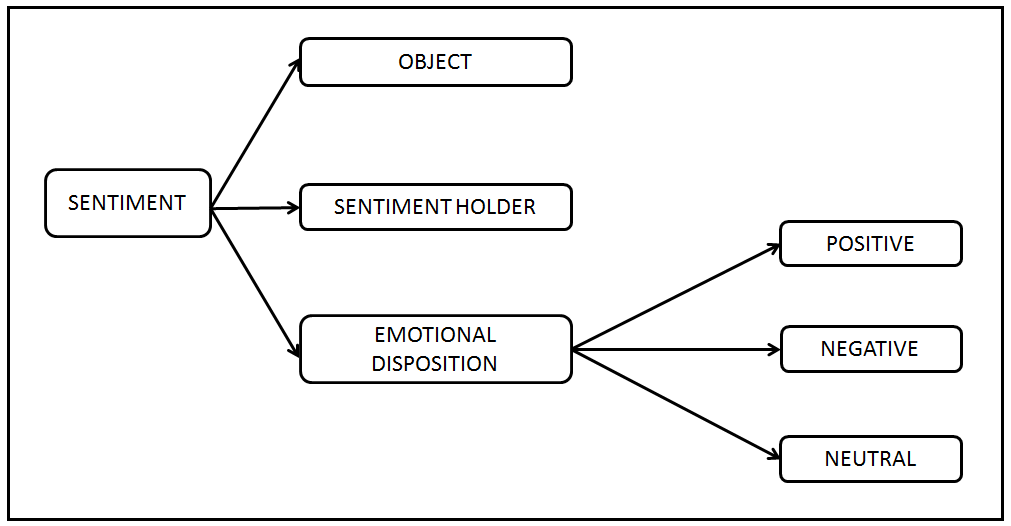}
		\caption{Schematic structure of sentiments \citep{munezero2014they}, which further divide into sentiment holder, emotional disposition and object of the review. The emotion is short-term, while the mood is long-term disposition.}
		\label{Data5}        
	\end{center}
\end{figure}

A schematic representation of opinion and sentiment are given in Figure ~\ref{Data5}.

SA has raised a growing interest in financial and political forecasting, e-health, e-tourism, and dialogue systems. The authors \citep{gallege2016towards} proposed trust-based ranking and the recommendation tool to improve online software services recommendation. This system enhances the existing recommendation system (content-based and collaborative filtering based) algorithm by considering the external attributes. The proposed system result was evaluated on the Amazon marketplace review dataset and showed a better ranking. 

SA used many libraries such as TextBlob \footnote{https://textblob.readthedocs.io/en/dev/} and naive Bayes to classify the content based on polarity score and subjectivity. In \citep{fang2015sentiment}, the authors proposed a sentiment polarity categorization process, which was the fundamental problem of sentiment earlier. The amazon product reviews dataset's experimental results achieved an F1 score of 0.8 and .0.73 for sentence-level and review-level categorization, respectively. The polarity shift problem is one of the challenges in sentiment analysis to predict user reviews. Xia et al. \citep{xia2015dual} proposed dual sentiment analysis model to address the polarity shift problem. The model trained for sentiment-reversed review for both training and testing. The result of the multi-domain and Chinese datasets showed the effectiveness of the model. 

In \citep{dasgupta2015sentiment}, the authors proposed a map reducing paradigm to collect the user's data from Facebook to understand brand reviews. They refined their approach through the iterative process of data pre-processing. In \citep{anto2016product}, the authors proposed an automatic feedback technique based on Twitter data. Different classifiers like SVM, Naive Bayes, and maximum entropy are used on Twitter comments. Out of these classifiers, SVM-based performance was the highest. In \citep{tan2011random}, the authors proposed a random walk algorithm for domain-oriented sentiment lexicon based on utilizing sentiment words and documents from both the old and target domains. The proposed algorithm reflects four kinds of relationships (words to documents, words to words, documents to words, and documents to documents) between words and documents. Experimental results indicate improvements in identifying the polarities of sentiment words. Day et al. \citep{day2016deep} presented that analytical methods use deep learning in financial news sources to forecast stock price trends. The authors found that financial news media sources can reveal investment information. Sentiment analysis aims to classify text in positive and negative polarity scores useful in quantifying different affective states of a user \citep{poria2017review}. Cambria et al. \citep{cambria2017sentiment} have developed an NLP approach that leverages both data and theory-driven methods to understand natural language.\\
Many approaches have simple categorization problems; however, sentiment analysis is a big suitcase problem requiring multiple polarity detection tasks. NLP problems divide into three layers: syntactic, semantics, and pragmatics. Each layer has a different subtask to process each layer's text output as input for the next layer. In \citep{poria2016deeper}, the authors have developed a pre-trained model for extracting emotion, sentiment, and personality features from sarcastic tweets using CNN. Experiments were conducted on a dataset consisting of both sarcastic and non-sarcastic tweets. Results computed on three datasets with $F_1$ scores of 87\%, 92.32\%, and 93.30\%, respectively.

\subsection{Sentiment in Text}
Text is mainly an important medium to express the user's state of mind in reviews and comments on the internet. It was the primary mode of communication in early 1990 when e-commerce company amazon was the first company to do business online. In 1992, the authors \citep{hearst1992direction} proposed an approach based on the sentence's directionality. The approach is based on the semantic orientation of the sentence to determine the directionality of the text. Another researcher \citep{sack1994computation} whose theory is based on the information's subjective point of view.
In the early 2000s, many researchers \citep{das2001yahoo, morinaga2002mining, pang2002thumbs, tong2001operational, turney2003measuring, wiebe2000learning} worked on sentiments analysis and opinions mining. Nasukawa et al. \citep{nasukawa2003sentiment} showed the high precision result on customer reviews and news articles available over web pages. They classified the specific subjects from a document in positive or negative polarity. This paper's result rise in interest to other researchers in this domain. The influential 2008 review of Pang and Lee \citep{pang2008opinion} covers techniques and approaches that promise to directly enable opinion-oriented information-seeking systems on benchmark datasets in recent research. Here, we discuss sentiment analysis in NLP, including its different methods, such as supervised and unsupervised.
\subsubsection{Supervised Approach}
It is based on the annotated dataset (labeled data) to build a prediction model. This approach builds a feature vector of the text, either aspect or word frequency, then the model learns (training) on the dataset and gives prediction for unseen data in testing. The first paper \citep{wiebe1999development} used this approach to classify the text as subjective or objective on the gold standard dataset. It achieves 81.5\% accuracy on the probabilistic classifier. There are different approaches in machine learning, like the supervised and unsupervised approaches. The supervised approach was used on stock trading \citep{das2007yahoo} domain to find the sentiment analysis. Further development focus on user comments available on the E-commerce site.
Many machine learning algorithms like SVM, Naive Bayes, and linear regression solve the problem related to a different domain. SVM was the most suitable model for product reviews in supervised sentiment analysis.  

The authors \citep{devlin2018bert} proposed BERT (bidirectional encoder representations from transformers) model designed to pre-trained deep bidirectional from the unlabelled text by jointly conditioning on both left and right context. Bidirectional means that BERT learns information from both the left and the right side of a token’s context during the training phase. The model is implemented on eleven NLP tasks such as GLUE, MultiNLI, SWAG, SQuAD v1.1, etc., and shown the impressive state-of-the-art result in the SA field. Unlike recent language representation models \citep{peters2018deep, radford2018improving}, this model is used for a wide range of tasks like question answering and language inference.  
\subsubsection{Unsupervised Learning Approach}
In this approach, the labeled data is not present, allowing an estimation based on expert knowledge.  The most popular method in unsupervised learning is cluster analysis to find the data's hidden pattern. In sentiment analysis, the lexicon plays an essential role in classifying the text into positive, negative, and neutral depending on the lexicon method, a combination of words or phrases. The most popular lexicon is General Inquirer \citep{stone1997thematic}, it is a corpus of positive and negative terms.

The aim to improve sentence-level classification, recently few methods are performing well such as SentiStrength \citep{thelwall2010sentiment}, Valence Aware Dictionary and sEntiment Reasoner (VADER) \citep{hutto2014vader}, and Umigon \citep{levallois2013sentiment}. VADER is a lexicon and rule-based sentiment analysis method used to find the sentiment of the reviews available on different social media platforms. The reviews are shared by a user from different age groups and gender, so these reviews are not available in the form that can be directly processed by any method. It converted into a normalized form after pre-processing of these reviews. VADER evaluates the words and their context on pre-processed reviews based on a predefined dictionary with many words (sentiment lexicon) and their corresponding numeric score. The method produces four metrics for each review; the first three are positive, neutral, and negative. The last metric is a compound score used to identify these reviews' sentiment. VADER is popular among other methods to analyze social media posts with slang, emoticons, and acronyms.  \\

Du et al. \citep{du2018text} proposed an attention mechanism for news categories in Chinese (NLPCC201) and English (REV1-v) datasets. The experimental result shows that this mechanism is beneficial to assign a score for keywords. The keywords that have a higher score in the corpus mean that these keywords are more important to the dataset than non-key words, so it improves the classifier's accuracy compared to recurrent neural network \citep{mikolov2010recurrent} and long short term memory \citep{hochreiter1997long}. This mechanism showed an effective result in many research papers \citep{zadeh2018multi,yang2016hierarchical,yan2019stat} for a different domain such as document classification (yelp reviews, IMDB reviews, Yahoo answers, and amazon reviews), understand human communication (language, vision, and acoustic modality), and video captioning, etc. \\

\subsubsection{Word Embedding}
The development of deep learning techniques in sentiment analysis shows promising results in most real-world problems. Word embedding is the dominant approach in NLP problems compare to one-hot encoding. If the words are present in the vocabulary in one-hot encoding, then assign one else zero. The issue in one hot encoding is a computational issue. When you increase your vocabulary by size n, the feature size vector also increases by length n, requiring more computational time to train the model. A word embedding is a learned representation for text data where words or phrases with the same meaning have a similar representation mapped further either in vector or real numbers. The strategy typically includes a mathematic concept from a high-dimensional vector space to a lower-dimensional vector space.
The vectors encoding is related to linguistic regularities and patterns, each dimension related to the word's feature. 
The learning of word embedding is done by neural network  \citep{bengio2003neural} from the text. \\
The most common word embedding system is word2vec, in which the words related to each other, like king-queen and man-women, are represented in the vector space near each other.  
The word2vec model approach is based on two models i.e. continuous bag-of-words \citep{mikolov2013efficient} and skip-gram model \citep{mikolov2013distributed}. Another frequent word embedding technique is Glove Vector \citep{pennington2014glove} (GloVe), which utilizes both global statistics and local statistics to train word vector fast and scalable.
The word2vec captures local statistics to do works well on analogy tasks. The author\citep{araque2017enhancing} proposed ensemble techniques that were the combination of word embeddings and a linear algorithm on seven public datasets extracted from the microblogging and movie reviews domain. This paper showed that word embedding techniques enhance the proposed model's performance and work well in a smaller dataset. The deep learning algorithm does not perform very well when the dataset is small because it requires a large amount of data to train the model, so the word embedding algorithm is used in this case. Pre-trained word embedding is used to solve many research problems \citep{ren2016topic, tang2014learning, giatsoglou2017sentiment}.  

\subsubsection{Others Techniques}
Qazi et al. \citep{qazi2017assessing} proposed assessing users' opinions on multiple topics like a social get-together, promoting efforts, and item inclinations. This study aims to find the users' expectations and satisfaction at the post-purchase stage. They surveyed a questionnaire comprising seven sections, and the data was collected through LinkedIn and the university mail servers. The authors utilized a disconfirmation hypothesis, a set of seven theories, confirmatory factor analysis, and primary conditions to break down the users' information and assess the model. The model's consequences demonstrated that regular, comparative, and interesting assessments positively raise users' desires. The author presumed that a wide range of sentiments is a rich wellspring of data that at last influences the customer loyalty level.
Wang et al. \citep{wang2018sentirelated} proposed a SentiRelated algorithm to fill the gap between different domains.
The traditional supervised classification algorithm is performed well for a given domain but does not work well on different domains. The SentiRelated algorithm is based on the Sentiment Related Index to improve the model's performance when tested on other domains.
This algorithm was validated on two datasets with different domains such as a computer, Education, Hotel, Movie, Music, and Book reviews and showed 80\% accuracy for short texts.  Social media platforms like Twitter are trending to become a common platform for exchanging raw data and online text, providing a vast platform for sentiment analysis.\\ The author proposed \citep{pandey2017twitter} a novel metaheuristic method based on Cuckoo Search and K-means (called CSK). It enlightens the clustering-based methods for analysing Twitter tweets to find the user's viewpoints and the sentiment pertained while making such a tweet. The method proposed outlines to find the optimum cluster-heads from the Twitter dataset's sentimental contents. The model tested its efficacy on various Twitter datasets and then compared it with the existing methods such as particle swarm optimization, differential evolution, cuckoo search, improved cuckoo search, etc. This research work performed a basis for designing a system that quickly provides conclusive reviews on any social issues.\\ 
The authors \citep{saif2016contextual} discussed an approach where a publicized tweets from the Twitter site are processed and classified based on their polarity. In this paper, a new tool is called "SENTICIRCLE" is used a lexicon-based approach. The word's semantics are extracted from its co-occurrence pattern, and the strength is updated in the lexicon accordingly. The basic idea of the approach is that the group of word accompanying it decides the semantic of the word in any text. The force of movement from the static word sentiment orientation approach to this contextual approach derived from the dictum "YOU SHALL KNOW THE WORD BY THE COMPANY IT KEEPS." It is different from the traditional lexicon, where the words are given fixed static semantics regardless of the context.\\

\subsubsection{Microblogging Data of Non-English Text}
Many studies have been conducted in sentiment analysis on English texts, while other languages have less attention than Arabic, Hindi, Bangla, etc. Many researchers have worked on SA in different languages after the rise of Web 2.0. The author in  \citep{al2019comprehensive} introduced an overview on Arabic assessment analysis \citep{badarneh2018fine, al2017application, socher2013recursive} in which they examined various tools and applications pertinent to it. The study additionally included both corpus-based and dictionary-based approaches for different datasets. Microblogs like tweets are trending rapidly for online users to share their experiences and opinions daily.
In contrast to the online reviews and blogs, these microblogs contain very dispersed and incomplete data. Unlike English-based microblogs, Chinese microblogs such as Sina Weibo have less sentiment analysis. The reason being that Chinese textual analysis is more challenging than English as its grammar of expression is different. The same length of Chinese sentences may contain more data than English, and the separation of words in those texts is relatively obscure. In totality, textually analyzing Chinese blogs has three primary research goals: First, the new words mining and their sentiment inference; second, how to extract other media modules and third, establish a hierarchical sentiment detection method based on Sina Weibo linguistics.
The authors \citep{wang2014sentiment} proposed three primary goals for the analysis of these Chinese microblogs. They visualize the sentiment analysis's result, depicting the relationship between social network sentiments and real-life events.\\
The researchers already working on Chinese microblogs tend to analyze the topic focussing on a single attribute while neglecting others. The model design is multilevel in single-level features keeping all the aspects under consideration.
Chen et al. \citep{chen2017improving} used to extract the text's sentiment using sentence-level sentiment analysis, but unlike other traditional approaches where the same technique was used in all types of sentences. The sentences are classified into three groups based upon their opinion targets. There are other ways to classify the sentences that have been previously used in other research papers. For example, The sentence can be subjective or objective based upon the subjectivity of the sentences. The subjective sentences express the opinions, while objective sentences implicate opinions or sentiments. The opinionated targets focused on the primary sentence classification. This opinion target can be any entity on which opinion is expressed. These opinionated sentences can give an opinion without mentioning the target on three different types of sentences: non-target, one-target, and multi-target. The Bi-LSTM and CNN deep learning approaches were used to classify the sentences and extract the text's syntactic and semantic features.

\subsection{Sentiment in Speech}
Analysis of speech in search of emotional and affective cues has a comparably long tradition \citep{dellaert1996recognizing}. This paper proposed statistical pattern recognition techniques to classify 1000 utterances according to their emotional content. Meanwhile, several kinds of literature have been established, including a range of recent surveys in emotions and affect in speech \citep{schuller2011recognising}. However, targeting sentiment explicitly exclusively from spoken utterances is a comparably new field than text-based sentiment analysis. Focusing on the acoustic side of spoken language, the border between sentiment and emotion analysis is often fragile, as discussed in \citep{crouch2012sentiment}. Mairesse et al. \citep{mairesse2012can} focused on pitch-related features and observed that pitch contains information on sentiment without textual cues. The authors collected short-spoken reviews from 84 speakers, and the result outperformed a majority class baseline. This paper attracted other researchers to explore this area to solve real-world problems.
The authors \citep{elmadany2018arsas} created Arabic Speech Act and Sentiment (ArSAS) dataset. The dataset consisted of 21,064 tweets annotated for two tasks: speech act recognition and SA. Further, the tweets are annotated for four different sentiment categories: positive, negative, neutral, and mixed.\\
Ahmed et al. \citep{ahmed2016agent} showed the sentiment in phone calls by first using speech recognition to extract the text in the call and then use typical text-based SA techniques. The goal was to measure agent productivity in call centers. 
       
\subsection{Image based Sentiment Analysis}
Vision-based emotion recognition \citep{zeng2008survey, sariyanidi2014automatic,campos2017pixels} is a relatively recent area of research. Users share millions of images and videos over social media platforms like Twitter, Tumblr, Flickr, and Instagram. These are the most popular sites where celebrities from sports, entertainment, and politics field share information in images. In the image-based sentiment analysis, opinions depict in the form of cartoons or memes. In most cases, the information conveyed through images is more effective compared to other modalities. 
Multiple techniques and algorithms such as SVM, naive Bayes, maximum entropy, and deep learning have been proposed in the image-based sentiment area to get significant results. 
The first work introduced by \citep{mikels2005emotional} to classify the images into positive and negative. The author showed that there is a strong correlation between sentiment images and their visual content. \\
Further, the SentiWordNet lexicon \citep{ohana2009sentiment} was used to find the text's numerical scores associated with the image. This lexicon is used in WorldNet databases to identify the positive and negative sentiment of the word. Emotions are difficult to identify and pin down when discussing the state of the emotion that differentiates from other emotional states. To find the scientific approach regarding the emotional state of the human being. The database of different photos was collected and validated against the specific emotional response of the viewers. This database is called International Affective Picture System (IAPS). Mikels et al. \citep{mikels2005emotional} studied eight emotion output categories: awe, anger, amusement, contentment, excitement, disgust, sadness, and fear. The author showed that each emotional state is different as different emotions have other cognitive and behavioral consequences. This paper adds some new dimensions of data for IAPS.

\subsection{Multimodal Sentiment Analysis}
Multimodal sentiment analysis \citep{lakomkin2019incorporating} performs sentiment analysis from multiple data such as audio, video, and text. It is the new dimension of traditional text-based sentiment analysis. 
Poria et al. \citep{poria2016fusing} proposed a new multimodal sentiment analysis methodology, which outperformed state of the art by more than 20\%. The proposed system used feature-based fusion techniques on text, visual and audio data from the youtube dataset. Different classifiers such as Naive Bayes, SVM, and extreme learning machines are implemented on the youtube dataset. The results showed that the extreme learning classifier is better than other classifiers. Extreme learning classifiers have single layer or multiple layers of hidden nodes. In most cases, the weights of the hidden nodes are learned in a single step so the overall processing time to classifying the result is less. 

Kumar et al. \citep{kumar2019fusion} proposed a multimodal rating prediction framework for products to improve customer satisfaction. The forty participants were participated in this study to collect EEG data of the product. The text's reviews from the product are processed through NLP techniques. The customer's rating from EEG and the product's reviews fused through optimization techniques. The experiment result showed that the ABC optimization approach was better than the unimodal scheme.  
There are many languages other than English, where researchers are working to predict the sentiment \citep{xu2019chinese,yang2020sentiment,behera2021co, zhao2020user,williams2018dnn}. In \citep{khasawneh2015arabic}, the authors proposed a hybrid approach on the Arabic dataset (Text and audio). Two machine learning approaches were used on this dataset to find the polarity. The bagging and boosting algorithms were used to enhance the proposed system further. 
The summary of the reviewed articles is as shown in Table \ref{tab:smry}.

\begin{table}[!t]
	\centering
	\tiny
	\caption{Summary of the publication's details included author, approach, data set, and accuracy.}
	\label{tab:smry}
\begin{tabular}{|c|c|c|c|}
\hline
\textbf{Author \& Year}         & \textbf{Approach}                                                                                                & \textbf{Dataset}                                                                                  & \textbf{Accuracy (\%)} \\ \hline
Hearst et al. {[}14{]}, 1992    & Cognitive Linguistics                                                                                            & User's Query                                                                                      & -                      \\ \hline
Wiebet et al. {[}21{]}, 1999    & Probabilistic Classifier                                                                                         & Gold-standard                                                                                     & -                      \\ \hline
Nasukawa et al. {[}22{]}, 2003  & Sentiment Lexicon                                                                                                & \begin{tabular}[c]{@{}c@{}}Camera reviews, \\ news articles\end{tabular}                          & 75-95                  \\ \hline
Pang et al. {[}23{]}, 2008      & \begin{tabular}[c]{@{}c@{}}Supervised and Unsupervised \\ approach (Survey paper)\end{tabular}                   & -                                                                                                 & -                      \\ \hline
Tan et al. {[}11{]}, 2011       & \begin{tabular}[c]{@{}c@{}}Domain-Oriented \\ sentiment  lexicon\end{tabular}                                    & \begin{tabular}[c]{@{}c@{}}Electronics reviews, \\ Stock reviews \\ and Hotel review\end{tabular} & 82.9                   \\ \hline
Fang et al. {[}9{]}, 2015       & \begin{tabular}[c]{@{}c@{}}sentiment polarity \\ categorization process\end{tabular}                             & \begin{tabular}[c]{@{}c@{}}amazon product \\ reviews\end{tabular}                                 & 80                    \\ \hline
Gallege et al. {[}8{]}, 2016    & \begin{tabular}[c]{@{}c@{}}Trust-based ranking \\ and the recommendation\end{tabular}                            & \begin{tabular}[c]{@{}c@{}}Amazon \\ marketplace \\ review\end{tabular}                           & -                      \\ \hline
Xia et al. {[}10{]}, 2015       & \begin{tabular}[c]{@{}c@{}}Naive Bayes, linear SVM, \\ logistic regression\end{tabular}                          & \begin{tabular}[c]{@{}c@{}}Multi-Domain and \\ Chinese dataset\end{tabular}                       & 90                    \\ \hline
Khasawneh et al. {[}12{]}, 2015 & Bagging and Boosting                                                                                             & \begin{tabular}[c]{@{}c@{}}1500 Arabic \\ comments and \\ Twitter reviews\end{tabular}            & -                      \\ \hline
Campos et al. {[}65{]}, 2017    & CNN                                                                                                              & Twitter images                                                                                    & -                      \\ \hline
Poria et al.  {[}13{]}, 2016    & ELM classifier                                                                                                   & YouTube Dataset                                                                                   & -                      \\ \hline
Cambria et al. {[}1{]}, 2017    & Top-Down and Bottom-Up                                                                                           & \begin{tabular}[c]{@{}c@{}}Penn Treebank, \\ LIWC\end{tabular}                                    & -                      \\ \hline
Kumar et al. {[}4{]}, 2019      & ABC optimization                                                                                                 & \begin{tabular}[c]{@{}c@{}}EEG data and \\ product reviews\end{tabular}                           & -                      \\ \hline
Qazi et al. {[}47{]}, 2017      & \begin{tabular}[c]{@{}c@{}}Expectancy disconfirmation \\ theory and Confirmatory \\ factor analysis\end{tabular} & \begin{tabular}[c]{@{}c@{}}LinkedIn and \\ the university mail \\ servers' groups\end{tabular}    & -                      \\ \hline
Wang et al. {[}48{]}, 2018      & SentiRelated                                                                                                     & \begin{tabular}[c]{@{}c@{}}Raw Data and \\ Douban Data\end{tabular}                               & 80                     \\ \hline

Williams et al. {[}48{]}, 2018      & \begin{tabular}[c]{@{}c@{}} intermediate-level \\
feature fusion \end{tabular}                                                                                                     & \begin{tabular}[c]{@{}c@{}} MOSI dataset\end{tabular}                               & 74.0                     \\ \hline
Yang et al. {[}48{]}, 2020      & SLCABG                                                                                                     & \begin{tabular}[c]{@{}c@{}}book
reviews \\ collected from \\Dangdang dataset\end{tabular}                               & \begin{tabular}[c]{@{}c@{}} Accuracy 93.5\\ Precision 93\\ Recall 93.6\\ F1 93.3\end{tabular}                     \\ \hline
Xu et al. {[}48{]}, 2019      & Seninfo+TF-IDF                                                                                                    & \begin{tabular}[c]{@{}c@{}}15000 hotel \\ comment texts \end{tabular}                               & \begin{tabular}[c]{@{}c@{}} Precision 91.54\\ Recall 92.82\\ F1 92.18\end{tabular}                     \\ \hline
 Lakomkin et al. {[}48{]}, 2019      & ASR model                                                                                                    & \begin{tabular}[c]{@{}c@{}} Multimodal Corpus of \\ Sentiment Intensity \end{tabular}                               & \begin{tabular}[c]{@{}c@{}} 73.6 \end{tabular}                     \\ \hline
  Guo et al. {[}48{]}, 2022      & \begin{tabular}[c]{@{}c@{}} CNN-BiGRU-CTC +  \\ ERNIE-BiLSTM  \end{tabular}                                                                                                     & \begin{tabular}[c]{@{}c@{}} Aishell-1  \\ and NLPCC 2014 \end{tabular}                               & \begin{tabular}[c]{@{}c@{}} 94.5 \end{tabular}                     \\ \hline
  Kumar et al. {[}48{]}, 2022      & \begin{tabular}[c]{@{}c@{}} BiLSTM + GloVe   \end{tabular}                                                                                                     & \begin{tabular}[c]{@{}c@{}} IIT-R STSA \end{tabular}                               & \begin{tabular}[c]{@{}c@{}} 92.83 \end{tabular}                     \\ \hline
Tian et al. {[}48{]}, 2021      & \begin{tabular}[c]{@{}c@{}} BERT-LARGE + A-KVMN \\
with  second-order \\ word dependencies  \end{tabular}                                                                                                     & \begin{tabular}[c]{@{}c@{}} LAP14, REST14, \\ REST15, REST16 \\and Twitter \end{tabular}                               & \begin{tabular}[c]{@{}c@{}} 92.48 \end{tabular}                     \\ \hline
Behera et al. {[}48{]}, 2021      & \begin{tabular}[c]{@{}c@{}} Co-LSTM model \end{tabular}                                                                                                     & \begin{tabular}[c]{@{}c@{}} Movie review,\\ Airline dataset \\ Self driving car GOP \end{tabular}                               & \begin{tabular}[c]{@{}c@{}} 98.40 \end{tabular}                     \\ \hline
Zhao et al. {[}48{]}, 2020      & \begin{tabular}[c]{@{}c@{}}Attention-based \\LSTM model \end{tabular}                                                                                                     & \begin{tabular}[c]{@{}c@{}} Facebook corpus \\ containing user \\ personality tag \end{tabular}                               & \begin{tabular}[c]{@{}c@{}} Precision 57.95\\ Recall 65.78\\ F1 72.2 \end{tabular}                     \\ \hline
 Derakhshan et al. {[}48{]}, 2019      & \begin{tabular}[c]{@{}c@{}}LDA-POS model \end{tabular}                                                                                                     & \begin{tabular}[c]{@{}c@{}} English and  Persian \end{tabular}                               & \begin{tabular}[c]{@{}c@{}} English dataset \\ average  56.24\\ Persian dataset \\ average  55.33 \end{tabular}                     \\ \hline
\end{tabular}
\end{table}

There are many public datasets \citep{guo2022multiple,tian2021enhancing} available in SA. 
For different application such as text, images, audio and video, we use different datasets or create a dataset like Sanskrit dataset \citep{kumar2022zero}. The tools and software library is also depend on the multimodal data. Open CV is a open source library used in computer vision tasks for object dection, face recognition and image segmentation. NLTK is a python library used for understanding the text or speech. Below is some popular database as shown in Table \ref{tab:smry2}.

\begin{table}[]
	\centering
	\tiny
	\caption{List of most popular public datasets available in the sentiment analysis field in different languages and modalities. The details included the source and uses of the dataset like the movie reviews, product reviews, social media data, etc.}
	\label{tab:smry2}
\begin{tabular}{|c|c|c|c|}
\hline
\textbf{Research Paper}                                       & \textbf{Dataset}                                                  & \textbf{Use of this dataset}                                                                          & \textbf{Volume}                                                                                         \\ \hline
Bai et al. \citep{bai2011predicting}         & IMDB Movie                                                        & To analysis movie review                                                                              & 50,000 movie reviews                                                                                    \\ \hline
Li et al. \citep{li2013deriving}             & Twitter                                                           & \begin{tabular}[c]{@{}c@{}}To do the sentiment analysis \\ of tweets of different domain\end{tabular} & \begin{tabular}[c]{@{}c@{}}Google 640917, \\ Microsoft 161292 \\ and Sony 141529 \\ tweets\end{tabular} \\ \hline
Araque et al. \citep{araque2017enhancing}    & Sentiment140                                                      & \begin{tabular}[c]{@{}c@{}}Sentiment analysis of tweets \\ for a product or brand\end{tabular}        & 1.6 million tweets                                                                                      \\ \hline
Dredze et al. \citep{blitzer2007biographies} & \begin{tabular}[c]{@{}c@{}}Amazon Product \\ Reviews\end{tabular} & \begin{tabular}[c]{@{}c@{}}To classify user review in \\ positive and negative\end{tabular}           & 142.8 million review                                                                                    \\ \hline
Qian et al. \citep{lei2016rating}            & Restaurant Reviews                                                & \begin{tabular}[c]{@{}c@{}}To find the aspect based \\ sentiment analysis\end{tabular}                & \begin{tabular}[c]{@{}c@{}}3 million restaurant \\ reviews\end{tabular}                                 \\ \hline
Karyotis et al. \citep{karyotis2018fuzzy}    & Facebook                                                          & \begin{tabular}[c]{@{}c@{}}To do the sentiment analysis \\ of Facebook post\end{tabular}              & \begin{tabular}[c]{@{}c@{}}million Facebook users \\ and their posts\end{tabular}                       \\ \hline
Chen et al. \citep{xu2014visual}             & Flicker images                                                    & To classify the image                                                                                 & \begin{tabular}[c]{@{}c@{}}470 positive tweets \\ and 133 negative tweets, \\ Tumblr 1179\end{tabular}  \\ \hline
Yang et al. \citep{yang2018visual}           & IAPS, Instagram                                                   & visual sentiment prediction                                                                           & \begin{tabular}[c]{@{}c@{}}IAPS 395, \\ Instagram 23308\end{tabular}                                    \\ \hline
Yang et al. \citep{yang2019aspect}           & SemEval 2014                                                      & Aspect-based sentiment analysis                                                                       & \begin{tabular}[c]{@{}c@{}}Restaurants 3841, \\ Laptops 3845\end{tabular}                               \\ \hline
Zhang et al. \citep{zhang2018textual}        & SemEval 2016                                                      & Sentiment analysis track                                                                              & \begin{tabular}[c]{@{}c@{}}Positive 3094, \\ Negative 2043 \\ and Neutral 863 tweets\end{tabular}       \\ \hline
Schmitt et al. \citep{schmitt2018joint}      & SemEval 2017                                                      & \begin{tabular}[c]{@{}c@{}}Detecting sentiment, \\ humour, and truth\end{tabular}                     & 8000-10000 tweets                                                                                       \\ \hline
Joshi et al. \citep{joshi2010fall}      & Hindi Movie Reviews                                                      & \begin{tabular}[c]{@{}c@{}}Sentiment analysis in Hindi\end{tabular}                     & 250 Hindi
Movie Reviews                        \\ \hline
Xu et al. \citep{xu2019chinese}      & \begin{tabular}[c]{@{}c@{}}Reviews of hotel \\ clothes, fruit\\ digital etc \end{tabular}                                             & \begin{tabular}[c]{@{}c@{}}Chinese Text Sentiment \\ Analysis\end{tabular}                     &  2,50000 reviews                         \\ 
\hline
Stappen et al. \citep{stappen2021multimodal}      & \begin{tabular}[c]{@{}c@{}}MuSe-CAR \end{tabular}                                             & \begin{tabular}[c]{@{}c@{}}The Multimodal \\ Sentiment Analysis  \\ in Car Reviews\end{tabular}                     &  15 GB Audio, Video, Text                          \\ 
\hline
Latif et al. \citep{latif2018cross}      & \begin{tabular}[c]{@{}c@{}}URDU-Dataset \end{tabular}                                             & \begin{tabular}[c]{@{}c@{}}4 emotions: angry, \\ happy, neutral, and sad. \end{tabular}                     &  0.072 GB Audio                         \\ 
\hline
Duville et al. \citep{duville2021mexican}      & \begin{tabular}[c]{@{}c@{}} MESD\end{tabular}                                             & \begin{tabular}[c]{@{}c@{}}6 emotions provides\\ single-word utterances\\  for anger, disgust, fear\\ happiness, neutral, and sadness. \end{tabular}                     &  0,097 GB Audio                         \\ 
\hline
\end{tabular}
\end{table}

\section{Usage and Application of Sentiment Analysis}
\label{sec:application}
This section covers the wide range of applications of sentiment analysis in various emerging areas.   

\subsection{Reviews from E-commerce and Microblogging Sites}
We have an extensive collection of data sets available on almost everything over the internet. It includes user comments, reviews, feedback on various topics, opinions drawn using surveys, products on e-commerce websites \citep{haque2018sentiment}, customer services \citep{kang2014based}, and recently Twitter data in the form of tweets on ongoing COVID-19 \citep{abd2020top,manguri2020twitter,alamoodi2021sentiment,chakraborty2020sentiment,naseem2021covidsenti} pandemic. Therefore, there is a severe demand for a system based on sentiment analysis that can extract sentiments about a particular product, item, or service.   It will help us to automate the user feedback or customer rating for the given product, services, etc. This would help to improve the product and offered services and eventually serve both the buyer and seller's requirements.

\subsection{Business Intelligence}
Nowadays, consumers are getting more intelligent \citep{sreesurya2020hypex}, quality-conscious, and technical savvy; therefore, they tend to seek out the reviews and ratings of online products and services before buying them. Many companies like Uber \citep{baj2017sentiment}, Oyo \citep{shanmugam2020twitter}, and zomato \citep{gupta2021sentiment} use digital transformation models to take feedback from the customer. The online customer opinion decides the success or failure of their offered services and products.    
The companies demand to extract sentiment from the online user reviews to enhance their offered products and services. It also helps companies to launch their new products and services in the new market for target customers. Therefore, It is evident that sentiment analysis plays a vital role in getting the customer and competition insights, which help companies make corrective and preventive actions to sustain and grow their businesses in the digital era.   

\subsection{Global Financial Market}
\begin{figure}[!h!t!b]
	\begin{center}
		\includegraphics[width=1\linewidth]{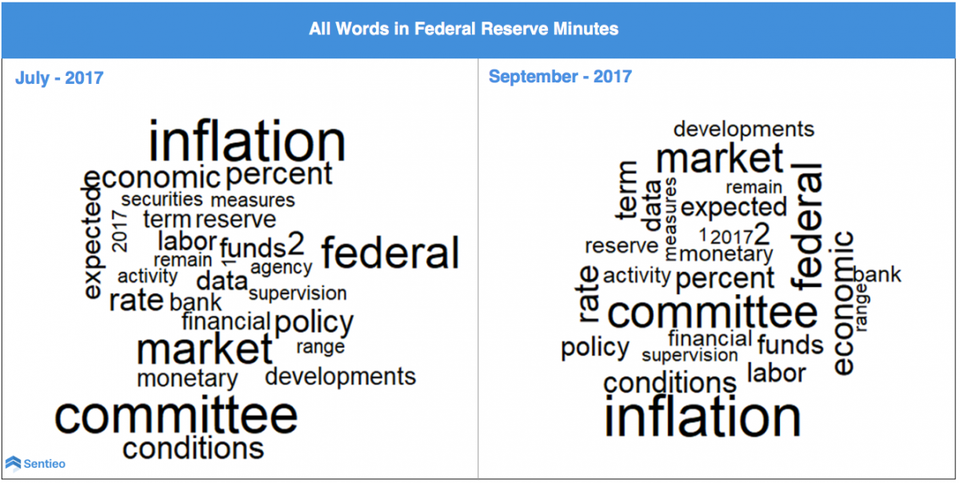}
		\caption{The federal open market committee (FOMC) controls the monetary policy of the central bank. The FOMC's statement lexical frequency list (most popular word in the report) of July and September 2017.}
		\label{Data10}        
	\end{center}
\end{figure}

\begin{figure}[!h!t!b]
	\begin{center}
		\includegraphics[width=1\linewidth]{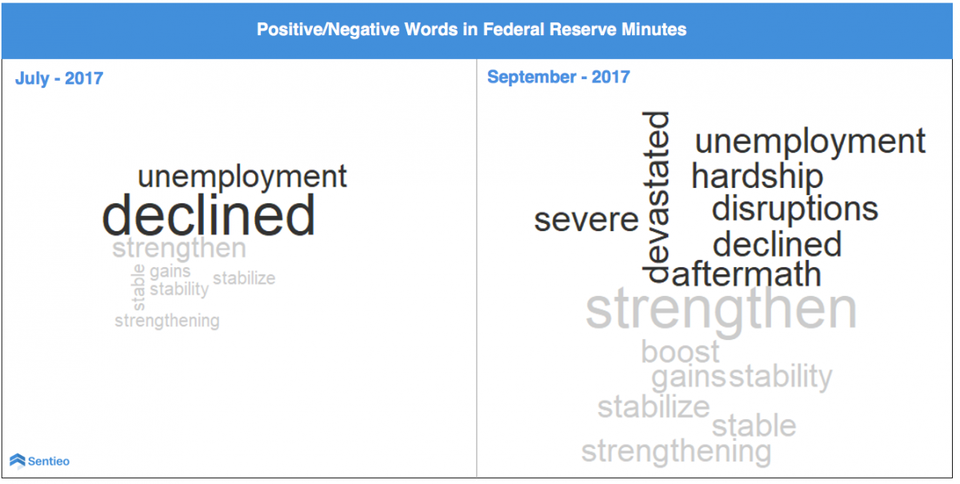}
		\caption{In the list of positive and negative words in the FOMC's statement of July and September 2017, the words in bold font denote the negative word while the other indicates the positive word.}
		\label{Data11}        
	\end{center}
\end{figure}

\begin{figure}[!h!t!b]
	\begin{center}
		\includegraphics[width=1\linewidth]{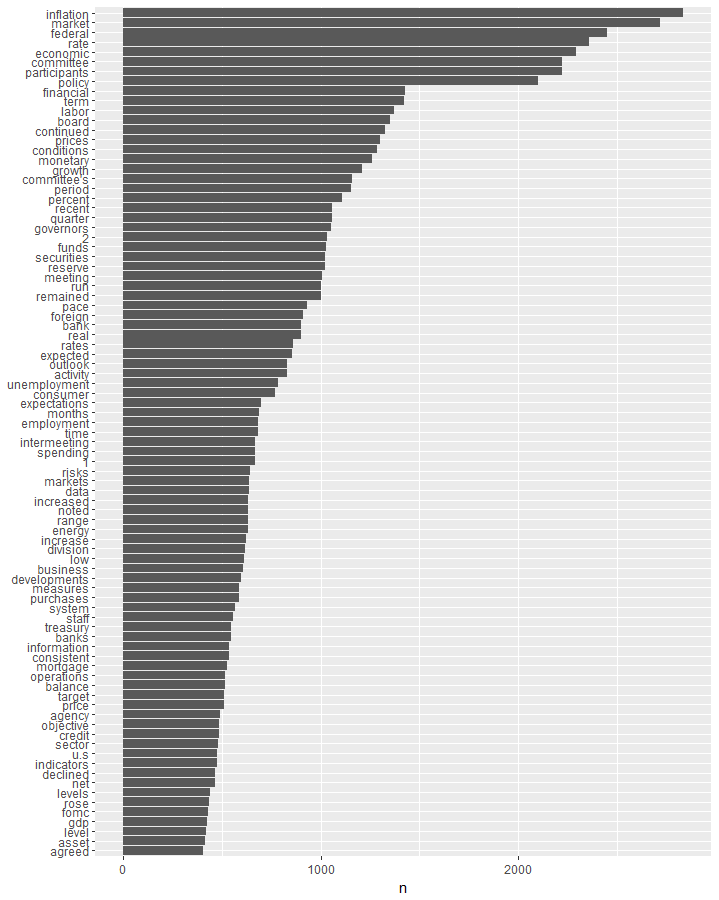}
		\caption{Most commonly used words in the FOMC statements since 2012, where n is the occurrence of words. In this chart, the most frequent word has shown from top to bottom; the top word in the chart has a higher occurrence than the last word.}
		\label{Data12}        
	\end{center}
\end{figure}

 SA is also helpful in the share market and the Federal open market committee (FOMC \footnote{https://www.federalreserve.gov/monetarypolicy/fomccalendars.htm}) statement \citep{tadle2022fomc,bhandari2022sentiment,doh2021you} to extract the meaningful information for traders through which they understand the global financial markets \footnote{https://www.forbes.com/sites/alapshah/2017/09/22/sentiment-analysis-of-fomc-statements-reveals-a-more-hawkish-fed/?sh=4079d5a8632e}. Some interesting trends reveal in Figure \ref{Data10}, Figure \ref{Data11} and Figure \ref{Data12} through sentiment analysis of FOMC statements.

\subsection{Applications in Smart Homes}
Smart homes are an emerging technology, and in the near future, the entire home will be more secure and better connected with other home appliances. The people would control and manage any part of the house using smart wearable devices such as apple watch, intelligent assistant devices such as Alexa \citep{bogdan2021practical,gao2018alexa}, Google Home \citep{sanchez2021using, park2018perception}, etc.   Recently there has been a lot of research going on in the Internet of Things (IoT) and the SA. The SA also found its way in IoT, e.g., the connected home using smart devices such as smart bulbs, smart music devices could alter its ambiance to create a calming and comfortable environment based on the sentiment or emotion of the user.

\subsection{Detection of Hate Speech}
Bigotry speech \citep{gitari2015lexicon,badjatiya2017deep,rodriguez2019automatic}  is used to express repugnance towards a specifically intended community, group, or person that can cause a dangerous situation to the victim. It can also be used to demean or offend particular community members or groups on any social media. SA based detection system would help the social media companies such as Twitter \citep{jiang2019detecting}, Instagram \citep{naf2019sentiment}, etc., instant messaging companies such as WhatsApp \citep{deb2020framework},  Telegram  and local enforcement and government to suppress hate speech and fake news towards a specific person, sex, religion, race, country, etc. which in turn improve their reputation and bring harmony in the community.   

\subsection{Emotion detection in suicide notes}
In modern society, suicides are rising rapidly in recent times; it is critical to find a faster way to fine-grained emotion detection \citep{ghosh2022deep, desmet2013emotion,prasad2018sentiment} and anxiety in online posts, microblogging text in the form of tweets by these troubled individuals. The SA-based detection and analysis system may help to detect such tendencies upfront and prevent suicides.

\subsection{Stress Detection}
On the flip side of excessive competition, improving the living style in a fast-moving world, people typically face many changes from their work environment, eating habits, etc. The body reacts to these stress changes, influencing an individual's emotional, mental, and physical health. The SA based detection system may help to detect stress symptoms \citep{wang2013depression,jung2017ontology} upfront and prevent any adverse impact due to this.\\



\section{Challenges and Perspectives}
\label{sec:challenges}
The SA is a particularly challenging task for human behaviors and subjective sentiments. Below are a few of the challenges-

\subsection{Recognizing Subjective Parts of The Phrase}
The English language can sometimes be tricky. Homonyms, or multiple-meaning words, have the same spelling and usually sound alike but have different meanings. Subjective parts in the phrase or sentence epitomize sentiment-related content. The Homonyms in the phrase might be treated as subjective in one case or objective in some other. It brings it challenging to identify the subjective portions of the phrase. For example:
1. The new lamp had good light for reading. 
2. Magnesium is a light metal. 
The word light is used to mean a particular quality or type of light in the first phase, whereas the light word objectively means having a relatively low density in the second phrase.
 Users share views or opinions over the internet on different social media platforms. Different age groups and gender share information or opinion in their way, recent study \citep{kumar2020exploring} prove that older people people share their opinion in a better way instead of young ones. 

\subsection{Dependence on The Domains}
The same phrase might have different interpretations in different domains in which it is being used. For Example, the word 'unpredictable' is positive in entertainment and theater,  etc., but if the same word is used in the context of an automobile's break, it has a negative opinion. Still, this is challenging to identify the domain from which any word is related correctly. Different pre-trained word embedding corpus domains such as IMDB movie reviews corpus and customer reviews dataset classify the sentence correctly. This challenge is still not solved completely, and researchers are continuously working on this problem.

\subsection{Detection of Sarcasm in The Phrase}
Sarcastic sentences express a negative opinion about a person or thing using positive words in unique. Often, people use it to say the opposite of what's true to make someone look or feel foolish. For Example: -" Good perfume. You must marinate in it for long". The sentence has only positive words, but it expresses a negative sentiment.

\subsection{Dependence on The Order}
Discourse Structure analysis is essential for opinion mining and sentiment analysis. For Example, A is better than B conveys the exact opposite opinion from B is better than A. For finding SA for these kind of sentence is quite challenging. 

\subsection{Idioms}
ML programs are designed so that they don’t understand a figure of speech. For example, language such as "not my cup of tea" will disrupt the algorithm because it understands the things literally. When any user uses idioms in a comment or review, the sentence interpretation is not correctly map by the algorithm. The situation is even more difficult if the comment is multilingual.

\subsection{Multilingual sentiment analysis}
User share their opinion in different languages like Hinglish which is the combination of Hindi and English. Every language has its own lemmatizer, POS tagger and grammatical constructs so ML or Deep learning algorithm understand the context and classify the comment in positive and negative. The real challenges is that we can not translate multiple language into one base language. Usually in micro-blogging or chatting, user share their feeling in multilingual.                

\begin{figure}[!h!t!b]
	\begin{center}
		\includegraphics[width=0.9\linewidth]{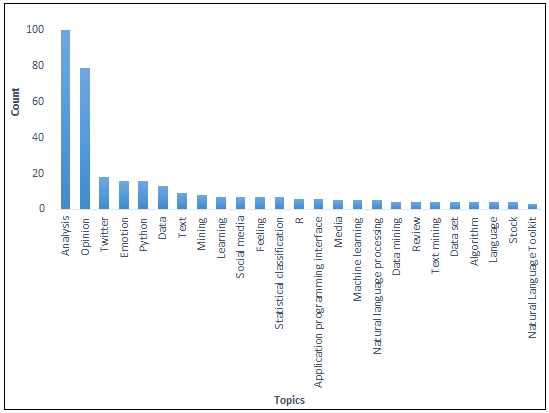}
		\caption{The top 25 topics search by the user worldwide from 2004 to 2020 in the sentiment analysis field. The most frequent topics search by the users is analysis and opinion.}
		\label{Data13}        
	\end{center}
\end{figure}

\begin{figure}[!h!t!b]
	\begin{center}
		\includegraphics[width=.9\linewidth]{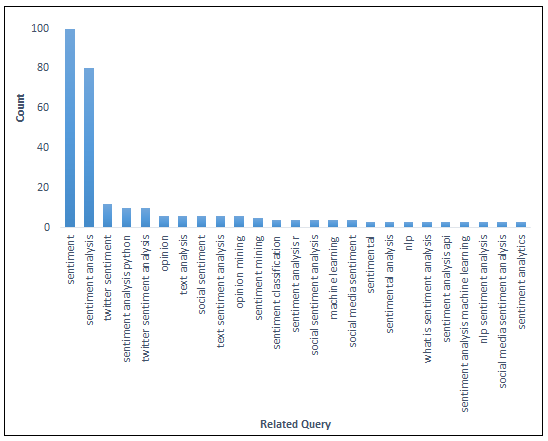}
		\caption{The most frequently searched query worldwide from 2004 to 2020 in the sentiment analysis field. The sentiment, sentiment analysis, and Twitter sentiment are the top three search queries worldwide.}
		\label{Data14}        
	\end{center}
\end{figure}

 
Despite different challenges in sentiment analysis still, it is an emerging field among customers for decision-making. Figure \ref{Data13} and \ref{Data14} presents users' most popular topic and query search from 2004 to 2020.\\

\section{Discussion Towards ML and DL Techniques on Sentiment Analysis Field}
\label{sec:Discussion}
In the last decade, the paradigm shifted from machine learning to deep learning techniques. In-text data, the context problem is a big challenge to understand the sentence's meaning through the ML algorithm correctly. This problem solves through pre-trained word embedding and the VADER approach even we have a smaller training dataset. However, the pre-trained word embedding corpus was trained on the google news dataset (100 billion words) and IMDB movie dataset. It shows a good result when the data are related to the pre-trained corpus domain; otherwise, it will not predict the result as expected.
The BERT model is the start of the art model in NLP. It uses the bidirectional training of the input, which provides a more profound sense of the language context. However, it is very compute-intensive and takes time to predict the result. ML techniques are also predicted good results, depending on the dataset and nature of data. DL techniques predict a good outcome for a large dataset.

The present study covered different domains like text, speech, image, and video to analyze sentiment. In all domains, the start of the art algorithms and papers were discussed in the study. 

\section{Conclusion and Future Work}
\label{sec:conclude}
With the advancement of technology in machine learning and deep learning, the SA plays a vital role in analyzing data available on the internet in text, image, and speech. The SA is computationally identifying the polarity of text into a positive, negative, and neutral review. In this survey paper, we have investigated the history of the SA and its impact on the research community from the years 2000 to current trends. In the last five years, most articles are related to social media such as Facebook, Instagram, and Twitter. Most articles are related to the application area of health, restaurant, travel, spam, and politics.
We have also included the top-cited paper and discuss the research challenges and perspectives suitable for new researchers who want to start research in the ML, NLP, and SA fields. We also cover in detail about global finacial market (FOMC), different languages like Sanskrit, Hindi, Arabic, Chinese etc and modalities in which many authors used SA. In future work, the SA is combined with network traffic to detect fake opinion or news, which creates a serious problem, resulting in mob violence. The method to do the SA will also improve with the continuous advancement of the NLP and ML fields.

\section*{Compliance with ethical standards}
\textbf{Conflict of interest} The authors declared that they have no conflicts of
interest to this work.

\bibliographystyle{spbasic}       

\bibliography{egbib}
%
%

\end{document}